Sebastien GUINARD, Univ. Grenoble Alpes, CEA, DRT F-38000 Grenoble

# Prompt Readiness Levels (PRL): a maturity scale and scoring framework for production grade prompt assets.

Why prompt engineering needs a TRL like framework.

## Abstract.

Prompt engineering has become a production critical component of generative AI systems. However, organizations still lack a shared, auditable method to qualify prompt assets against operational objectives, safety constraints, and compliance requirements. This paper introduces Prompt Readiness Levels (PRL), a nine level maturity scale inspired by TRL, and the Prompt Readiness Score (PRS), a multidimensional scoring method with gating thresholds designed to prevent weak link failure modes. PRL/PRS provide an original, structured and methodological framework for governing prompt assets specification, testing, traceability, security evaluation, and deployment readiness enabling valuation of prompt engineering through reproducible qualification decisions across teams and industries.

## Introduction.

From TRL to prompt asset: why maturity scale, creating a common language, matters.

The deployment of generative AI systems in production faces a central challenge: mastering the quality of outputs. Unlike traditional software based on strictly deterministic inputs and processing, Large Language Models (LLMs) produce probabilistic responses dependent on their training on the one hand and the provided prompt and its integration into operations on the other hand. These introduce variability that impacts the reliability, stability, and reproducibility of behaviors in production.

If, just a few years ago, prompts were simple text inputs, they now become a central control surface for large language model based systems deployed in production connected with users, tools, agents and external data sources. In these systems, prompts are no longer informal instructions written once, integrated with limited engineering rigor and forgotten. They shape system behavior, safety properties, operational cost, and regulatory exposure. In practice, prompts have become engineering artifacts, yet they are rarely treated as such. Despite their growing importance, there is currently no shared way to answer a simple question: is this prompt ready to be deployed in a production or regulated environment?

The analogy with aerospace systems is striking: a poorly designed or integrated prompt can cause an interface failure and compromise the outcome. Inspired by the original work of Sadin and Mankins who developed and codified the nine level Technology Readiness Levels (TRL) scale in 1974 and 1995 for NASA[1,2], we propose the PRL: Prompt Readiness Levels and the PRS: Prompt Readiness Score to provide a robust basis for the development, audit, compliance demonstration, and potential monetization of prompt engineering within industrial settings. Indeed, TRL like approaches have successfully reduced ambiguity in communication between engineers, managers, and regulators by replacing subjective claims with explicit readiness levels. Prompt engineering faces a similar coordination problem. While the artifacts differ, the underlying need is the same: a stage gated language that connects technical evidence to deployment decisions.

Thus, this work defines the foundational concepts and reference definitions of PRL: Prompt Readiness Levels and PRS: Prompt Readiness Score as an integrated framework to characterize a prompt asset, being an auditable and governable engineering artifact rather than a mere text prompt, understood as a versioned and auditable package that bundles (i) a prompt specification (template, instructions, examples), (ii) an explicit interface (input/output schema), (iii) an execution context (model binding, inference parameters, tool/retrieval policies), (iv) an assurance package (test suite and acceptance criteria), (v) traceable evidence (evaluation reports and known limitations), and (vi) governance metadata (ownership, approvals, IP/licensing), all under a unique version identifier. It is intended to serve as a reference framework for subsequent developments and intellectual property considerations.

PRL/PRS is published as a canonical specification (PRL/PRS v1.0, 2026). Unless stated otherwise, this document is released under CC BY 4.0 (Attribution): reuse, implementation and adaptations, including commercial use, are permitted provided that the original work is credited and the version is indicated. Recommended attribution: *“Prompt Readiness Levels (PRL) & Prompt Readiness Score (PRS); Sébastien Guinard - v1.0 (2026) - CC BY 4.0 - canonical source: arXiv (forthcoming; identifier pending)”*. Adaptations are welcome; however, “PRL/PRS” refers to the canonical specification published by the author, and modified forks should clearly indicate deviations and use a distinct qualifier or name.

**State of the art and related work.**

Why prompts and their engineering are hard to qualify and what PRL/PRS changes in practice.

Technology Readiness Levels (TRL) have inspired metrics for software or software subsets (System Readiness Levels), applied in industrial acquisitions or the development of critical systems. These approaches transpose the concept of progressive maturity to guide integration decisions and reduce failure risks, illustrating the value of stage gated readiness assessments beyond hardware[3].

Also, in our current artificial intelligence era, the NIST AI RMF[4] emphasizes the necessity of measuring the validity and reliability of an AI system within the framework of trustworthy AI, and integrating these measures into its subcategories and risk management practices. This framework thus constitutes a structuring tool for designing and deploying rigorous evaluations of AI system performance and behavior. At the same time, prompt engineering has rapidly evolved. Although advances such as chain of thought reasoning[5], programmatic optimization, and prompt orchestration[6] have shown notable improvements, the field still lacks a standardized, auditable framework and a universal maturity scale for explicit, robust, and reproducible qualification. Evaluating LLM based systems remains challenging because outputs are open ended and task dependent. And if holistic evaluation frameworks (HELM) broaden the metric space beyond accuracy, other recent work studies using strong LLMs as judges for ranking chat assistants highlights biases and reliability limitations of automated judging[7, 8, 9]. Moreover, security research further underscores prompt injection and jailbreaking as critical threats, including indirect prompt injection when external data is retrieved and interpreted as instructions. Thus, industry guidance lists prompt injection as a top risk, and academic studies and benchmarks provide taxonomies, empirical analyses, and standardized red teaming evaluation for such attacks and defenses[10, 11, 12, 13]. In parallel, emerging practices in prompt lifecycle management, prompt governance, and LLMOps tooling improve traceability and operational control, but they do not provide a shared readiness level scale that links conformance evidence to deployment and compliance decisions.

PRL and PRS address this gap by introducing a stage gated readiness scale and a multidimensional scoring method for prompt assets, enabling explicit, auditable qualification aligned with regulatory and compliance requirements, such as those defined by ISO/IEC[14]. PRL and PRS do not replace evaluation frameworks, security testing, or governance processes. Instead, they connect them since they provide a shared vocabulary for discussing prompt asset maturity, a way to link technical evidence to deployment decisions, a mechanism to compare prompt assets across teams and organizations and a bridge between engineering practice and compliance requirements. In this sense, PRL/PRS function as a coordination layer rather than a competing methodology.

**The PRL scale: methodological framework and design directions.**

The PRL scale is divided into 9 levels spread across three phases, marking the transition from semantic intention to a proven and potentially certified asset. So, rather than asking whether a prompt works, PRL asks whether a prompt asset has reached a given level of engineering maturity. Early levels focus on intent and specification. Intermediate levels emphasize robustness, repeatability, and security. Higher levels require traceability, operational evidence, and compliance readiness. Crucially, PRL is not a checklist of best practices. It is a stage gated model: a prompt asset cannot claim a higher level without satisfying the requirements of all lower levels.

PRL scale is detailed below according to the phase, name of the phase, level number, name of the level, description of the level and associated deliverables that constitute an evidence pack. Positioning on the PRL scale and the possibility of moving from one level to another are determined by the PRS and associated thresholds (detailed further).

- Phase I: Intent (concept and semantic genesis, exploration and viability)
  This phase validates the model's ability to understand and execute the fundamental task. It maps the human intent to model's capabilities.

  - PRL 1: initial semantic mapping.
    - Description: Identification of semantic needs, identification of the task, boundaries, and required knowledge base. Initial zero shot testing to validate that the model has the latent space capacity to handle the request.
    - Deliverables: Semantic functional and scope document; preliminary feasibility report.

  - PRL 2: structural architecture
    - Description: Development of the prompt's backbone, implementation of personas, delimiter strategy (XML/Markdown), and explicit constraint sets. Engineering of the output schema (JSON, Code, Prose).
    - Deliverables: Prompt structural blueprint; initial prompt version

  - PRL 3: behavioral logic validation and Proof of Concept (PoC)
    - Description: Empirical testing on a representative sample (relevant to the system or subsystem considered). Validation of "in context learning" (ICL) effectiveness on a representative sample. Verification of the "reasoning path" (Chain-of-Thought) and tone consistency.
    - Deliverables: Reasoning path analysis, behavioral validation report; qualitative success logs.

- Phase II: Stabilization (hardening and determinism)
  This phase transforms the semantic attempt into a robust, optimized component.

  - PRL 4: deterministic benchmarking and evaluation
    - Description: Testing on a gold dataset (ground truth), systematic evaluation using automated metrics (cosine similarity, exact match, or LLM as a judge). Introduction of few-shot prompting and calibration.
    - Deliverables: Performance baseline report, measures of precision, recall, and hallucination rates.

  - PRL 5: advanced optimization, logic & variance optimization
    - Description: Advanced pattern integration (Chain-of-Thought, ReAct, RICE…), Hyper-parameter tuning (Temperature, Top-P) to reduce semantic variance and eliminate hallucinations. Reduction of the token to performance ratio.
    - Deliverables: Optimization matrix, variance stability report, token efficiency analysis.

- PRL 6: systemic robustness and cross-model resilience
  - Description: Validation of model-agnosticism or fine-tuning for a specific model, stress-testing against noise (typos, ambiguous or degraded inputs…)
  - Deliverables: Robustness and resilience report and certificate, dependencies map.

- Phase III: Industrialization and compliance (qualification and operational production) This phase addresses system integration, security, and long-term lifecycle management into the production.

  - PRL 7: security and alignment
    - Description: Certification against prompt injections (Red teaming), jailbreaking, leakage and validation of ethical alignment and compliance (GDPR/EU AI Act…).
    - Deliverables: Adversarial test log report, safety & compliance audit.

  - PRL 8: orchestration and systemic integration
    - Description: Integration into an orchestrator (LangChain, Semantic Kernel…), implementation of prompt versioning (Git-like), automated unit tests and validation in a pre-production environment
    - Deliverables: Integration ready API package, automated CI/CD validation suite

  - PRL 9: production integration and certification
    - Description: Large or full scale deployment under LLMOps governance, real time and continuous monitoring of semantic drift and cost per inference, use of a version control system, establishment of a feedback loop for continuous improvement and full auditability
    - Deliverables: Final audit log, final asset valuation report, LLMOps governance protocol

**Figure1: Introducing the "Prompt Readiness Levels" (PRL) framework; overview.**

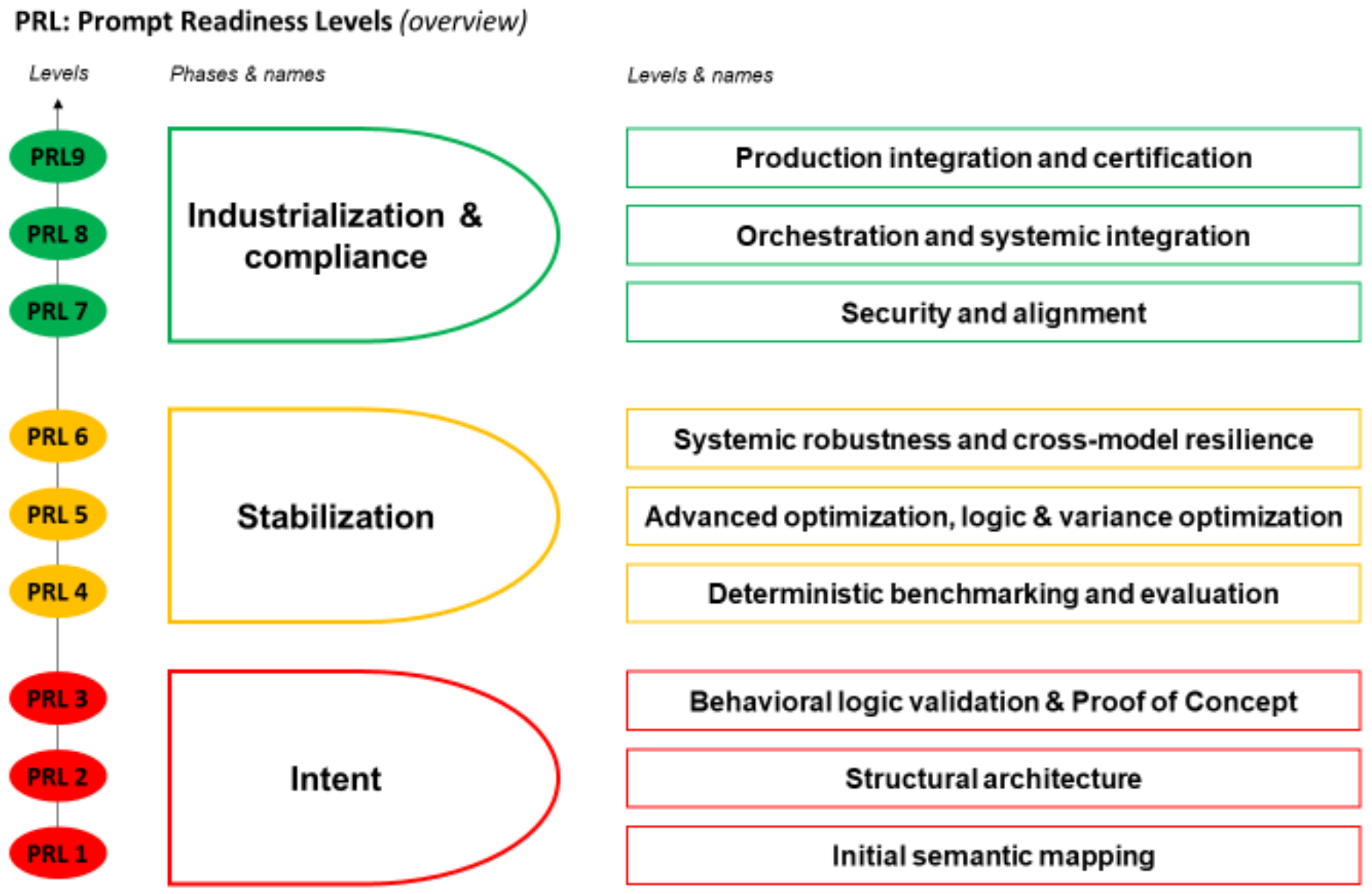

**The Prompt Readiness Score (PRS): a multidimensional metric.**

We distinguish a continuous score (PRS) from discrete readiness levels (PRL). The PRS aggregates multiple dimensions and the sensitivity to key operational and business considerations of the prompt asset. The PRS, generic at first glance, could be fine tuned to specific industries & business. The PRS is calculated based on five weighted dimensions, each representing a critical operation pillar.

- R - reliability & determinism: probability of consistent output vs. statistical variance.
- S - semantic integrity & resilience: robustness against linguistic drift, input noise, and context window saturation.
- C - compliance, safety & alignment: resistance to adversarial attacks (injections) and adherence to legal and, or, ethical frameworks.
- G - governance & asset traceability: level of documentation, version control, and intellectual property clear-path.
- O - operational efficiency and cost: token-optimization ratio, inference latency, and infrastructure compatibility.

PRS(**S**) is a weighted aggregation of dimension scores, $\boldsymbol{S} = (S_R, S_S, S_C, S_G, S_O))$, optionally penalized by instability across evaluation batches. PRL($\boldsymbol{S}$) is then defined as the highest level n such that PRS($\boldsymbol{S}$) meets the global threshold $\theta_n$ and all dimension scores meet the per level minima $\delta_{i,n}$ (no weak link gating). To prevent a good average from hiding a fatal flaw (for instance, a high performance prompt asset that is insecure), the PRS uses a veto function that acts as a no weak link.

PRS is expressed, for instance, according to the following conceptual equation and assessment function:

$$PRS(\boldsymbol{S}) = f(R, S, C, G, O)$$

$$PRS\ (\boldsymbol{S}) \in [0;100]$$

$$PRS(\boldsymbol{S}) = \max_{n \in [1,\ldots,9]} \left( \sum_{i\ \in\ \{R,S,C,G,O\}} (\omega_i\ .\ S_i\ .\ e^{-\lambda\sigma_i}\ .\ \Phi(i,n)) \right)$$

Where:

- $\omega_i$: the business priority coefficient, where $\sum \omega_i = 1$, $i\ \in \{R, S, C, G, O\}$
- $S_i$: the performance score of a prompt asset for dimension $i\ \in \{R, S, C, G, O\}$, $S_i\ \in$ [0;100]
- $\boldsymbol{S}$: the score vector obtained by a prompt asset; $\boldsymbol{S} = (S_R, S_S, S_C, S_G, S_O) \in [0;100]^5$
- $e^{-\lambda\sigma_i}$ is a linear penalty focused on the variance $\sigma_i$ to represent the instability of $S_i$ across test batches; λ being a scale factor to modulate the sensitivity to instability.

- $\Phi(i,n)$: the gatekeeper or veto function expressed in regards of the compliance vector $\boldsymbol{\delta_n}$, aggregating compliance thresholds $\delta_i, n$ that stands for the minimum acceptable score for a specific dimension at a given level n of the PRL scale, $n \in$ [0;9], $i \in \{R, S, C, G, O\}$:
  - $\Phi(i,n)$ = 1 if $\forall \mathrm{i}, S_i \geq \delta_{i,n}$
  - $\Phi(i,n)$= 0 if $\forall \mathrm{i}, S_i \geq \delta_{i,n}$
  - $\boldsymbol{\delta_n} = (\delta_{R,n}, \delta_{S,n}, \delta_{C,n}, \delta_{G,n}, \delta_{O,n}) \in$ [0;100][5]

Importantly, the PRS support the prompt asset positioning on the PRL scale, as:

$$PRL(\boldsymbol{S}) = \max\{n \in \{1,\ldots,9\} : \mathrm{PRS}(\mathbf{S}) \geq \theta_{\mathrm{n}} \wedge \forall \mathrm{i}, S_i \geq \delta_{i,n}\}$$

In other words, the highest level that passes global threshold and per dimension minima.

Then, tuples can be produced such as:

$$(PRL(\boldsymbol{S}) = n, \quad PRS(\boldsymbol{S}) = x, \quad \boldsymbol{S} = (S_R, S_S, S_C, S_G, S_O))$$
$$(\theta_{\mathrm{PRLn}} = y, \quad \boldsymbol{\delta_n} = (\delta_{R,n}, \delta_{S,n}, \delta_{C,n}, \delta_{G,n}, \delta_{O,n}))$$

Tagging developed prompt asset with decisive indicators such as:

$$(PRL(\boldsymbol{S}) = 4, \quad PRS(\boldsymbol{S}) = 68, \quad \boldsymbol{S} = (78, 83, 62, 70, 55))$$
$$(\theta_{\mathrm{PRL4}} = 65, \quad \boldsymbol{\delta_4} = (60, 60, 60, 50, 40))$$

**Discussion, limitations and open questions: toward a financial valuation of Prompt Readiness Levels (PRL) scale, Prompt Readiness Score (PRS) and prompt assets.**

The PRL: Prompt Readiness Levels scale and the PRS: Prompt Readiness Score secure industrial processes based on a prompt asset by structuring a controlled, governed repository and standardizing the production of technical, auditable, and traceable documentation, in compliance with critical and operational software engineering standards. They provide structured means to demonstrate prompt engineering compliance with ISO/IEC and other relevant standards, as well as with certifications mandated by current regulations[15]. Conformance claims to PRL shall be supported by an evidence package appropriate to the claimed level, including versioned prompt asset specifications, evaluation protocols, test results, traceability records, evidence pack and PRS. To make such claims easier to communicate, implementations may optionally use self-claimed conformance labels: implementations that follow the normative core of this specification (PRL/PRS vocabulary, gating logic, and minimum evidence per level) may describe themselves as “PRL-Conformant”. Implementations that reuse the framework with documented deviations may use “PRL-Compatible”. In both cases, the claim should cite the canonical source and the PRL/PRS version used.

PRL and PRS focus on prompt assets, not on models or entire systems. They should be used alongside broader AI risk management and governance frameworks. Also, open questions remain, including how best to calibrate readiness thresholds across domains and how to account for evolving models for instance. That’s why, to maximize adoption while preserving room for differentiation, we explicitly separate an open, normative core from optional proprietary extensions. The open specification defines the PRL/PRS vocabulary, the canonical gating logic (global thresholds and per dimension minima), and the minimum conformance evidence required at each level (test artifacts, traceability, security evaluation logs…), so that independent teams can implement, reproduce, and audit PRL claims consistently. In parallel, proprietary layers may legitimately exist as non normative profiles and implementations such as detailed assessment methodology, sector calibrated benchmarks, domain datasets, weighting schemes and thresholds tuned to business risk, automated evaluation tooling and operational integrations, provided they remain compatible with the open conformance rules. This two-layer approach enables PRL/PRS to evolve into a shared industry reference while allowing vendors and organizations to build differentiated, high value assessment products on top of a common standard.

**Referenced bibliography.**

1. J. Banke, Technology Readiness Levels Demystified, National Aeronautics and Space Administration (NASA), Washington, DC, USA, 2010.
2. J. C. Mankins, Technology Readiness Levels: A White Paper, National Aeronautics and Space Administration (NASA), 1995.
3. Office of the Under Secretary of Defense for Research and Engineering, Technology Readiness Assessment (TRA) Guidebook, U.S. Department of Defense, Washington, DC, USA, 2025.
4. National Institute of Standards and Technology (NIST), AI Risk Management Framework (AI RMF 1.0), NIST, Gaithersburg, MD, USA, 2023.
5. J. Wei, X. Wang, D. Schuurmans, et al., “Chain-of-Thought Prompting Elicits Reasoning in Large Language Models,” in Proceedings of the 36th Conference on Neural Information Processing Systems (NeurIPS), 2022.
6. O. Khattab, A. Singhvi, P. Maheshwari, et al., DSPy: Compiling Declarative Language Model Calls into Self-Improving Pipelines, Stanford University, 2023.
7. P. Liang, R. Bommasani, T. Lee, et al., Holistic Evaluation of Language Models, arXiv preprint arXiv:2211.09110, 2022.
8. L. Zheng, W.-L. Chiang, Y. Sheng, et al., Judging LLM-as-a-Judge with MT-Bench and Chatbot Arena, arXiv preprint arXiv:2306.05685, 2023.
9. A. Szymański, M. Szymański, and P. Przybyła, Limitations of the LLM-as-a-Judge Approach for Evaluating LLM Outputs in Expert Knowledge Tasks, 2024.
10. OWASP Foundation, OWASP Top 10 for Large Language Model Applications, Version 1.1, OWASP, 2025.
11. J. Greshake, S. Abdelnabi, S. Mishra, et al., Not What You’ve Signed Up For: Compromising Real-World LLM-Integrated Applications with Indirect Prompt Injection, arXiv preprint arXiv:2302.12173, 2023.
12. Y. Liu, H. Wen, J. Yu, et al., Jailbreaking ChatGPT via Prompt Engineering: An Empirical Study, arXiv preprint arXiv:2305.13860, 2023.
13. M. Mazeika, E. Denton, A. Artetxe, et al., HarmBench: A Standardized Evaluation Framework for Harmful Behavior in Large Language Models, arXiv preprint arXiv:2402.04249, 2024.
14. International Organization for Standardization / International Electrotechnical Commission, ISO/IEC 42001:2023 — Information Technology — Artificial Intelligence — Management System, ISO/IEC, Geneva, Switzerland, 2023.
15. European Commission, Regulation (EU) 2024 of the European Parliament and of the Council Laying Down Harmonised Rules on Artificial Intelligence (Artificial Intelligence Act), 2024.